\documentclass[11pt,a4paper]{article}
\usepackage[hyperref]{emnlp2020}
\usepackage{times}
\usepackage{latexsym}

\usepackage{microtype}

\usepackage{sec/packages}
\aclfinalcopy 

\title{Relation of the Relations: A New Paradigm of the \\
Relation Extraction Problem
}

\author{Zhijing Jin\thanks{ {} {}  Equal Contribution} \\
  AWS Shanghai AI Lab \\
  \texttt{zhijing.jin@connect.hku.hk} \\\And
  Yongyi Yang$^{*}$ \\
  Fudan University \\
  \texttt{17300240038@fudan.edu.cn} \\\AND
  Xipeng Qiu \\
  Fudan University \\
  \texttt{xpqiu@fudan.edu.cn} \\\And
  Zheng Zhang \\
  AWS Shanghai AI Lab \& NYU Shanghai \\
  \texttt{zhaz@amazon.com} 
  \\}

\date{}

\begin{document}
\maketitle

\begin{abstract}
In natural language, often multiple entities appear in the same text. However, most previous works in relation extraction (RE) limit the scope to identifying the relation between two entities at a time. Such an approach induces a quadratic computation time, and also overlooks the interdependency between multiple relations, namely the relation of relations (RoR). Due to the significance of RoR in existing datasets, we propose a new paradigm of RE that considers as a whole the predictions of all relations in the same context. Accordingly, we develop a data-driven approach that does not require hand-crafted rules but learns by itself the RoR, using Graph Neural Networks and a relation matrix transformer. Experiments show that our model outperforms the state-of-the-art approaches by +1.12\% on the ACE05 dataset and +2.55\% on SemEval 2018 Task 7.2, which is a substantial improvement on the two competitive benchmarks.\footnote{Our code is available at 
  \url{https://github.com/FFTYYY/RoR_relation_extraction}
}
\end{abstract}

\section{Introduction}\label{sec:intro}
\begin{figure}[!t]
    \centering
    \includegraphics[width= \columnwidth]{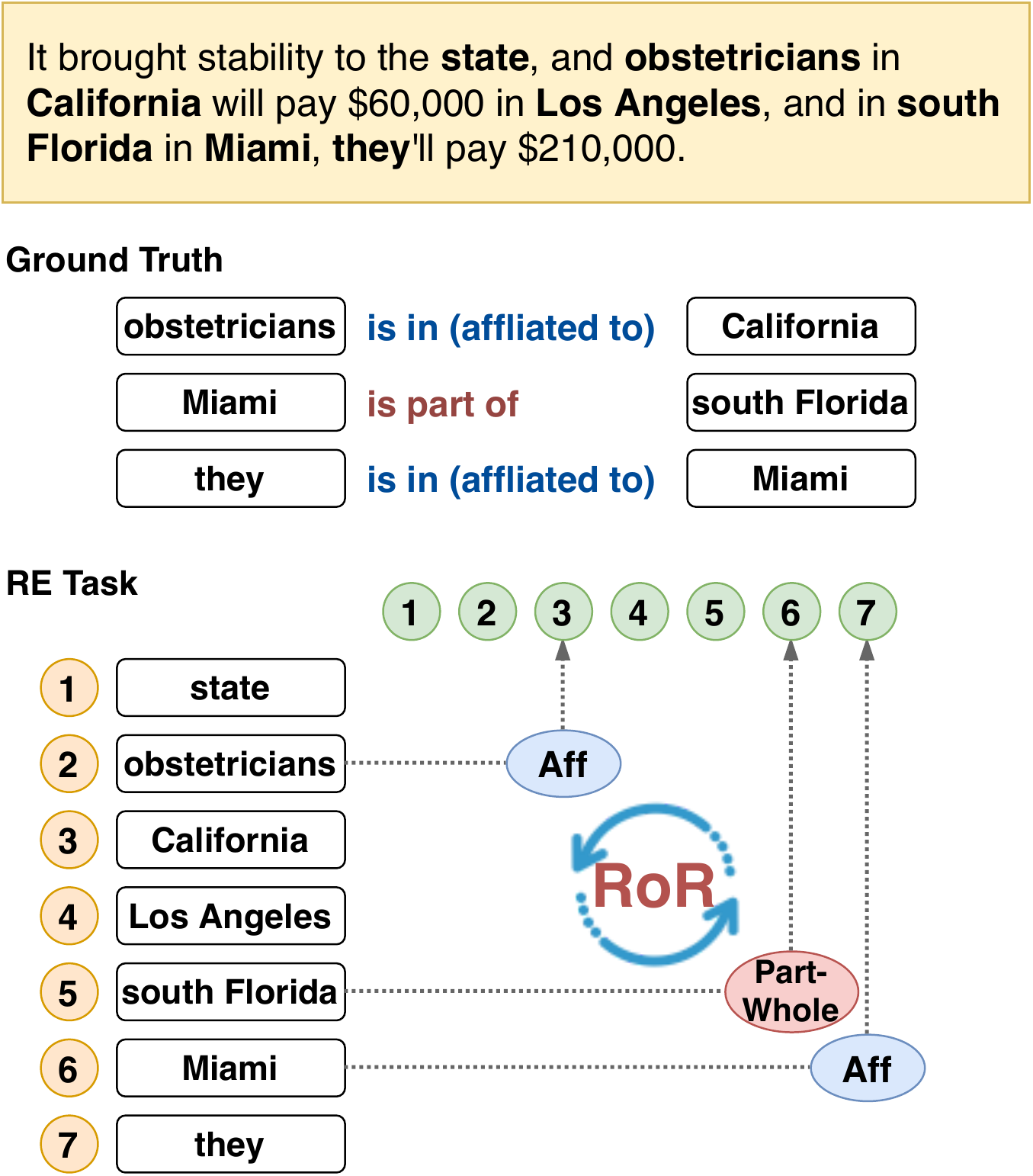}
    \caption{An example of the RE task.}
    \label{fig:intro-ex}
\end{figure}

Relation Extraction (RE) is the task to identify the relation of given entities, based on the text that they appear in. As a fundamental task of knowledge extraction from text, RE has become an active area of research in the past several decades~\cite{DBLP:conf/tipster/MillerCFRSSW98,DBLP:journals/jmlr/ZelenkoAR03,DBLP:conf/naacl/BunescuM05,DBLP:conf/acl/ZhouSZZ05,DBLP:conf/semeval/HendrickxKKNSPP10}. 

In natural language, most text includes multiple entities. For example, the sentence in Figure~\ref{fig:intro-ex} has seven entities. We find that 99.76\% data instances of the widely used ACE 05 dataset \cite{walker2006ace} have more than two entities, and there are 9.21 relations in each text on average. However, most previous research
has been confined to the simplified setting of only classifying the relation between every \textit{two} entities at a time \cite{DBLP:conf/coling/ZengLLZZ14,DBLP:conf/semeval/LuanOH18,DBLP:conf/acl/LiJ14,DBLP:conf/emnlp/GormleyYD15,DBLP:conf/acl/MiwaB16}. For the sentence in Figure~\ref{fig:intro-ex} with 7 entities, most previous approaches would perform 49 independent relation classification tasks (if including self-reflexive relations). It is not feasible to reduce this number of classification tasks because existing methods require explicit annotation of the entities in the input. For example, to predict the relation between the entity pair (obstetricians, California), the input needs to be transformed into ``...$\langle e1\rangle$ obstetricians $\langle \backslash e1\rangle$ in $\langle e2\rangle$ California $\langle \backslash e2\rangle$ will pay \$60,000 in Los Angeles ...''.

The problems exposed by such a previous paradigm is that it is not only inefficient, but also overlooks the interdependency among the multiple relations in one context. For example, in the 49 relations in Figure~\ref{fig:intro-ex}, if we already know the relationship (Miami, is part of, south Florida), where ``is part of'' is a relation defined on two objects, then it is very unlikely for \textit{Miami} to be in any other person-social relationship such as ``is the father of...''. 
Remember that, on average, there are 9.21 relations in each text in the ACE 05 dataset, for example, and each relation can provide information to other relations in the same text. We denote the frequently-appearing interdependency of the many relations in the same text as the ``relation of relations'' (RoR) phenomenon.

To capture RoR, we propose a new paradigm of RE by treating the predictions of all relations in the same text as a whole. Note that our work is distinct from \cite{DBLP:conf/acl/WangTYCWXGP19}, which still treats the relation of each entity pair as independent classification tasks, but saves the computation power at the cost of accuracy by encoding all entities in one pass.
Instead, our newly proposed paradigm is not about tradeoffs between computation cost and accuracy, but to increase the performance by capturing RoR.

In this paper, we first highlight the importance of RoR by identifying several types and their frequent occurrences in RE datasets in Section~\ref{sec:ror_stats}. We then propose a data-driven approach {without} hand-crafted rules, using Graph Neural Networks (GNNs) to model each relation as a node to learn the pair-wise dependency of every two relations, and then a matrix transformer to learn the correlations involving multiple relations or numerical correlations of the count of relations. We evaluate the model on two benchmark datasets, ACE 2005 (ACE05)~\cite{walker2006ace} and SemEval 2018 Task 7.2 (SemEval2018)~\cite{DBLP:conf/semeval/GaborBSQZC18}. Our system outperforms the previous state-of-the-art (SOTA) models by +1.12\% on ACE05 and +2.55\% on SemEval2018. The contributions of our paper are as follows:
\begin{itemize}
    \item We propose a new paradigm of RE, according to the frequent RoR phenomenon. It provides a new perspective for future research in RE.
    \item We develop a model to learn the interdependency of all relations in the same text, based on a GNN and matrix transformer.
    \item We validate the effectiveness of our model, which outperforms SOTA models by a clear margin on two benchmark datasets.
    \item We open-source our model and evaluation codes.
\end{itemize}

\section{New formulation of RE}
Most previous work formulates RE as multiple independent classification problems limited to two entities and the text: 
\begin{quote}
    Given two entity mentions $\bm{e}_1$ and $\bm{e}_2$, a text sequence $\bm{t}=\{w_1, w_2, \ldots, w_N\}$ involving $\bm{e}_1$ and $\bm{e}_2$, and a finite set relation types $\mathcal{R}$, the task is to predict the relation type between the two entities.
\end{quote}
Under this setting, RE can be solved by the well-researched sentence classification task.

\begin{figure}[h]
    \centering
    \includegraphics[width= \columnwidth]{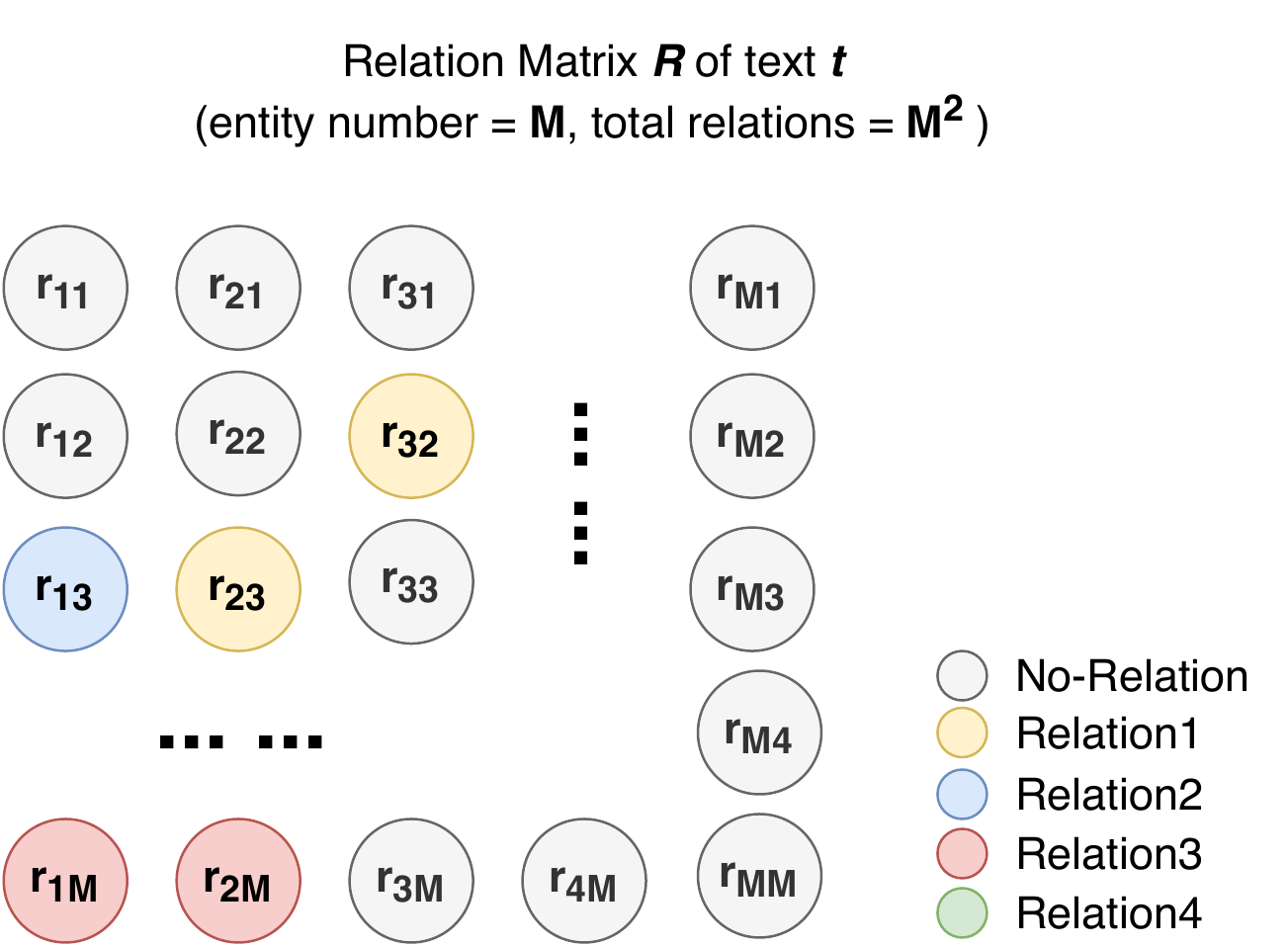}
    \caption{New formulation of RE that considers all $M$ entities in the same text.}
    \label{fig:problem_form}
\end{figure}

However, based on the motivations in Section~\ref{sec:intro}, we propose a new paradigm of RE:
\begin{quote}
Given~a~text~sequence~$\bm{t}=\{w_1, w_2, \ldots, w_N\}$ and all the entities $\bm{e}_1, \ldots, \bm{e}_M$ mentioned in $\bm{t}$, the model needs to predict the  relationship $r_{ij}$ between each two entities $(\bm{e}_i, \bm{e}_j)$, where $i, j \in \{1,\ldots, M\}$.
\end{quote}
We use a matrix $\bm{R} = (r_{ij}) \in \mathbb{R}^{M \times M}$ to represent all the relations of interest, as shown in Figure~\ref{fig:problem_form}.
\begin{table*}[!t]
  \centering
    \begin{tabular}{lll}
    \toprule
    \textbf{Relation} & \textbf{Arg0} & \textbf{Arg1} \\
    \midrule
    Person-Social (Per-Soc)
    (e.g., family) & PER & PER \\
    Part-Whole (e.g., city-and-country) & FAC, LOC, GPE, ORG & FAC, LOC, GPE, ORG \\
    Physical (Phys) (e.g., near) & PER, FAC, LOC, GPE & PER, FAC, LOC, GPE \\
    Organization-Affiliation (Org-Aff) (e.g., employment) & PER, ORG, GPE & ORG, GPE \\
    Agent-Artifact (Art) (e.g., owner-and-object) & PER, ORG, GPE & FAC \\
    General-Affiliation (Gen-Aff) (e.g., citizen) & PER & PER, LOC, GPE, ORG 
    \\
    \bottomrule
    \end{tabular}%
  \caption{Valid entity types for each relation. The entity type abbreviations refer to Person (PER), Facility (FAC), Location (LOC), Geo-PoliticalEntity (GPE), and Organization (ORG).}
  \label{tab:ror_type1a}
\end{table*}

\section{Statistical analysis of RoR}\label{sec:ror_stats}
We conduct our case study using the benchmark dataset ACE05 \cite{walker2006ace}. We will introduce two forms of RoR: (1) biRoR, which is only between two relations, and (2) multiRoR, which involves three or more relations. 

Note that the purpose of the following analyses is to demonstrate the importance of RoR. Our model introduced in the subsequent section will \textit{not} hand-craft such detailed rules but will learn RoR in a data-driven way.

\subsection{Data overview}

ACE05 \cite{walker2006ace} is the most widely used dataset for RE. Its text is extracted from a variety of sources, including news programs, newspapers, newswire reports, and audio transcripts. The 6 relation types are shown in Table~\ref{tab:ror_type1a}. There are 7 entity types valid for the relations: Facility (FAC), Geo-PoliticalEntity (GPE), Location (LOC), Organization (ORG), Person (PER), Vehicle (VEH), Weapon (WEA).

\subsection{BiRoR: Interdependency of two relations}\label{sec:ror_local}
\subsubsection{Entity type-constrained biRoR }\label{sec:corr1a}
We first introduce the simplest form of RoR, based on the entity types. Specifically, given the triple ($\bm{e}_1$, $\mathrm{RelType}_a$, $\bm{e}_2$) of two entities and their relation, we can infer whether $\bm{e}_1$ is unlikely to co-occur with a different relation type $\mathrm{RelType}_b$. 

To elucidate such constraint, we will make an example using the seven entity types in ACE05. As detailed in Table~\ref{tab:ror_type1a}, only certain types are allowed to be the arguments of the relations. Therefore, we can deduct 12 rules of \textit{incompatibility}. For example, 
the same entity cannot be both the arg0 of Per-Soc and the arg0 of Part-Whole, because an entity must be a person (PER) in order to satisfy the Per-Soc relationship, but Part-Whole \textit{cannot} involve PER. The full list of incompatibility rules are listed in Appendix~\ref{appd:corr1a}.

\subsubsection{Semantic-constrained biRoR}\label{sec:corr1b}
Another type of biRoR is constrained by the semantics of the relation. The intuition is that what a relation means can imply whether it can be shared or must be disjoint with another relation. For example, the Art relationship can describe a person (arg0) owning a facility (arg1), where the arg1 \textit{must} be a facility. If such relation already exists, the same facility cannot be involved in the Part-Whole relation with a city, because cities (e.g., Boston) cannot be a part of a facility, semantically. This kind of incompatibility is not a hard constraint by the entity type, but it is implied by the semantics of relations.

Semantics can also imply whether a relation should be symmetric. For example, Per-Soc is always symmetric because family and friends are commutative relations, whereas Org-Aff is always an asymmetric relation \cite{ACE05eval}. Hence, if a relation $r_{ij}=$ Per-Soc, then $r_{ji}=$ Per-Soc. And if $r_{ij}=$ Org-Aff, then $r_{ji}\neq$ Org-Aff.

\subsubsection{Empirical biRoR}\label{sec:corr1c}
\begin{figure}[ht]
    \centering
    \includegraphics[width= \columnwidth]{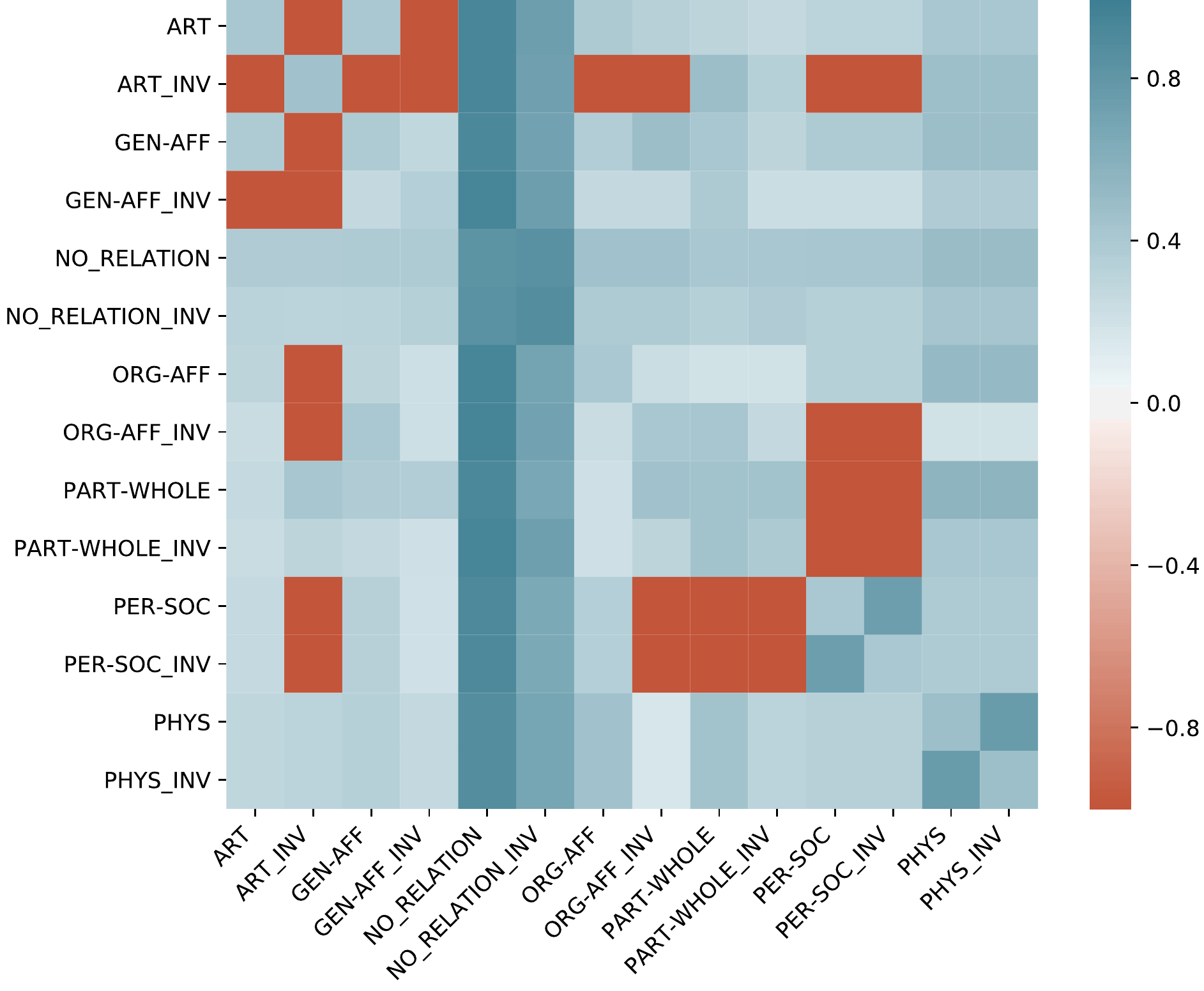}
    \caption{Correlation of every two relations. Each cell $(\mathrm{RelType}_i, \mathrm{RelType}_j)$ represents the conditional probability that an entity will have $\mathrm{RelType}_j$ given the existing $\mathrm{RelType}_i$. For better visual effect, the zero probability is converted to $-1$ (dark red), signaling that these two relations have 0\% co-occurrence.}
    \label{fig:corr2}
\end{figure}

To form a more direct understanding of biRoR, we calculate the correlation of every two relations in Figure~\ref{fig:corr2}. We can see that the incompatibility rules in Section~\ref{sec:corr1a} is proven by the red negative color, the symmetric property of Per-Soc and Phys in Section~\ref{sec:corr1b} is proven by the darker blue. 
There are also other correlations such as frequent 
co-occurrences of Part-Whole and Phys relations.

\subsection{MultiRoR: Correlation of 3+ relations}\label{sec:ror_nonlocal}

\subsubsection{Entity type-constrained multiRoR}
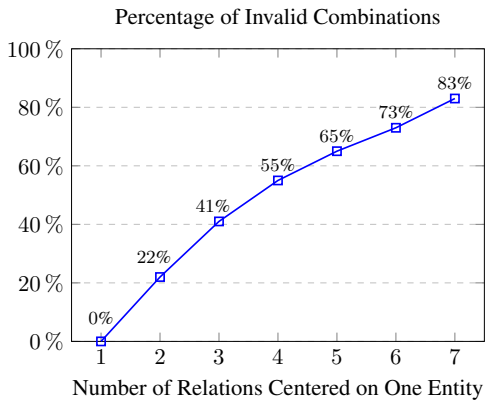
\begin{figure}[ht!]
  \centering
   \resizebox{0.85 \columnwidth}{!}{%
            \begin{tikzpicture}
            \pgfplotsset{
            scale only axis,
            legend style={at={(0,0)},anchor=south west},
        }
        
    \pgfplotsset{
percentage plot/.style={
nodes near coords align=vertical,
    yticklabel=\pgfmathprintnumber{\tick}\,$\%$,
    ymin=0,
    ymax=100,
    enlarge y limits={upper,value=0}
}
}
\begin{axis}[
    legend style={cells={align=left},nodes={scale=0.7, transform shape},
    legend cell align={left}},
    title={Percentage of Invalid Combinations},
    xlabel={Number of Relations Centered on One Entity},
    ylabel={},
    xmin=0.5, xmax=7.5,
    ymin=0, ymax=100,
    xtick={1,2,3,4,5,6,7},
    ytick={0,20,40,60,80,100},
    legend pos=north west,
    ymajorgrids=true,
    grid style=dashed,
    width=7cm,
    height=5cm,
    every axis plot/.append style={thick},
    percentage plot
]

\addplot[
    color=blue,
    mark=square,
    ]
    coordinates {
    (1,0)(2,22)(3,41)(4,55)(5,65)(6,73)(7,83)
    };
    \legend{}
    
\node [above,font=\small] at (axis cs:  1,  3) {\pgfmathprintnumber{0}\%};
\node [above,font=\small] at (axis cs:  1.9,  24) {\pgfmathprintnumber{22}\%};
\node [above,font=\small] at (axis cs:  2.9,  41.5) {\pgfmathprintnumber{41}\%};
\node [above,font=\small] at (axis cs:  4,  55.5) {\pgfmathprintnumber{55}\%};
\node [above,font=\small] at (axis cs:  5,  65.5) {\pgfmathprintnumber{65}\%};
\node [above,font=\small] at (axis cs:  6,  73.5) {\pgfmathprintnumber{73}\%};
\node [above,font=\small] at (axis cs:  7,  83.5) {\pgfmathprintnumber{83}\%};

\end{axis}
\end{tikzpicture}
}
\vspace{0px}
\caption{Percentage of invalid combinations over all possible set of relations of an entity, as the total number of non-empty relations centered on the entity increases.}
\label{fig:corr3_1}
\end{figure}

Other than biRoR, which only involves two relations, there are more complicated rules acting on multiple relations, namely multiRoR. The entity type-constrained multiRoR extends Section~\ref{sec:corr1a} from the incompatibility between two relations to among 3+ relations. For example, if we already know two relations of an entity, which is both the arg1 of the Org-Aff relation and the Phys relation, then this entity cannot be the arg1 of any Art relations (because this entity must be GPE). As the number of relations centered on one entity increases in Figure~\ref{fig:corr3_1}, the percentage of invalid combinations of multiple relations among all possible combinations will soon be over 50\%, and even reached 83\% when there are 7 relations.

\subsubsection{Numerically correlated multiRoR}

\begin{figure}[ht]
    \centering
    \includegraphics[width= \columnwidth]{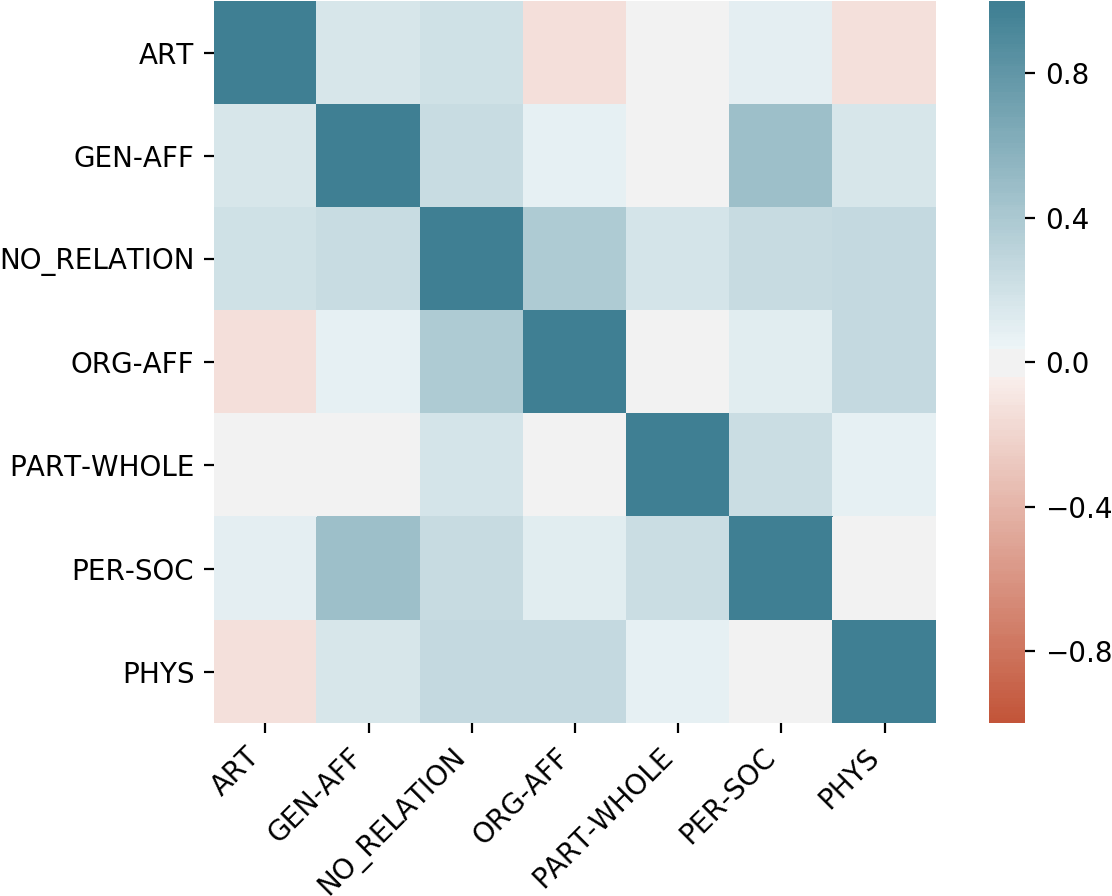}
    \caption{Numerical correlation of every two relation types. Each cell $(\mathrm{RelType}_i, \mathrm{RelType}_j)$ represents the linear correlation between the number of occurrences of $\mathrm{RelType}_i$ and that of $\mathrm{RelType}_j$.
    }
    \label{fig:corr3_3}
\end{figure}

We can also discover numerical correlations of multiple relations. From all relation matrices $\bm{R}$ in the dataset, we find that the number of occurrences of a specific type of relations can correlate with the number of another relation type. Note that it is counted as multiRoR, because the correlation is defined not between two single relations, but the total count of two relation types (each of which can include multiple occurrences). From the correlation plot in Figure~\ref{fig:corr3_3}, Per-Soc and Gen-Aff show a strong positive linear dependency, whereas Art and Org-Aff are negatively related by numbers.

\section{Method}
Based on the rich RoR phenomenon analyzed in Section~\ref{sec:ror_stats}, we aim to design a model that can mine these properties from data. A naive solution is to hand-craft many rules to impose every type of RoR, but it is not scalable when there are datasets of different features, or when there are some RoR that are difficult to be manually identified. Hence, we aim to design a model that has the capacity to learn RoR with no hand-crafting.

\begin{figure}[t]
    \centering
    \includegraphics[width= \columnwidth]{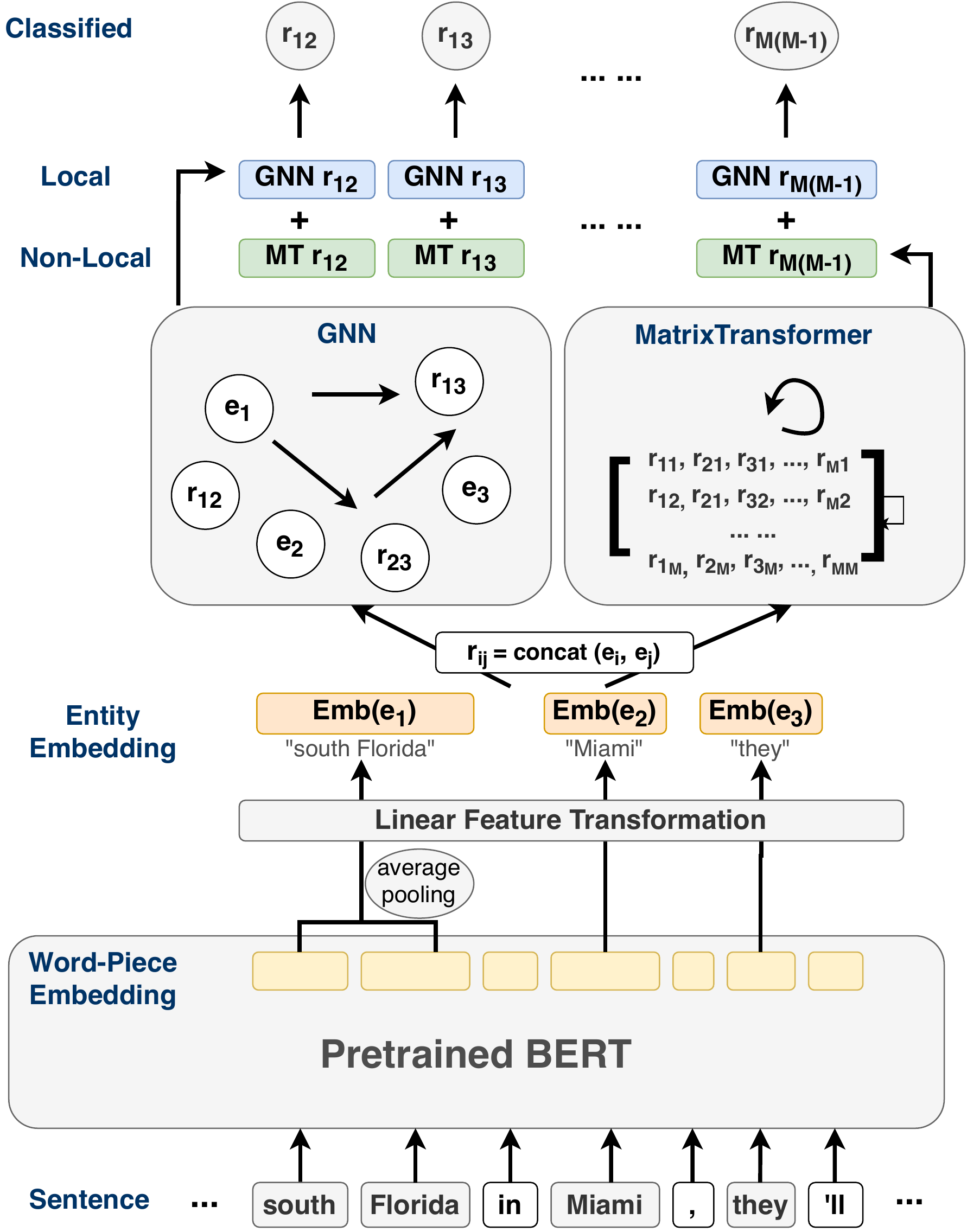}
    \caption{The model architecture of RoR.}
    \label{fig:archi}
\end{figure}

Our overall training strategy is in Figure~\ref{fig:archi}. In the following, we will elaborate on three key components: (1) initial embeddings of entities and relations, (2) the GNN-based biRoR learner, and (3) the matrix transformer that learns multiRoR.

\subsection{Initialization of entities and relations} \label{sec:method_init}
As a preparation step, we first obtain the embedding of each entity. We pass the text through a pretrained BERT model \cite{DBLP:conf/naacl/DevlinCLT19}, and obtain each entity representation by average pooling over its tokens' hidden states in the last layer of BERT. Note that our framework can easily adapt to other ways to retrieve pretrained embeddings, such as \cite{DBLP:conf/nips/YangDYCSL19,DBLP:journals/corr/abs-1907-11692}. 

Then, we obtain the initial embedding of each relation by applying an feedforward layer on the concatenated embeddings of the two involved entities.

\subsection{BiRoR Learner}\label{sec:method_local}
We use a GNN to learn the interdependency between every two relations, namely biRoR.
For each text sequence $\bm{t}$ with $M$ entities, we formulate a
graph $\mathcal{G} = (\mathcal{V},\mathcal{E})$, where all entities and relations form the nodes of the graph. Accordingly, we link every relation node to its two entity nodes.

We use a GNN to learn node-to-node interactions, and especially among the relation nodes. The first layer of GNN is initialized with the node embeddings (including all entity embeddings and relation embeddings) obtained from Section~\ref{sec:method_init}. In each layer $\ell$, it aggregates all hidden states passed from the neighbors to update its representation in the next layer $\ell +1$. More specifically, we have
\begin{align}
    h^{\ell+1}_u = \mathrm{FFN}\left(W^O\sum_{v \in {\mathcal{N}} (u)}\alpha_{u,v}h^{\ell}_v\right)
    ,
\end{align}
where $h_i^{\ell}$ is the hidden state of the $i$-th vertex in the $\ell$-th layer, $\mathrm{FFN}(\cdot)$ is a feed-forward network, $W^O$ is the weight matrix, ${\mathcal{N}} (u)$ is the set of neighbor nodes to the vertex $u$, and $\alpha_{u,v}$ is the attention weight that $u$ has to $v$. This attention weight $\alpha_{u,v}$ is obtained by
\begin{align*}
    \alpha_{u,v} = \frac{ \exp \left[W^{Q}h^{\ell}_v\left(W^{K}h^{\ell}_u\right)^T\right]}{ \sum_{v' \in {\mathcal{N}} (u)}\exp \left[W^{Q}h^{\ell}_{v'}\left(W^{K}h^{\ell}_u\right)^T\right]}
    ,
\end{align*}
where $W^{K}$ and $W^{Q}$ are the key and query weight matrices when calculating the attention.

\subsection{MultiRoR Learner} \label{sec:method_global}
The GNN introduced in Section~\ref{sec:method_local} has a strong ability to model node-to-node interaction, which corresponds to the biRoR, but it is not as strong when seizing the more complicated multiRoR as analyzed in Section~\ref{sec:ror_nonlocal}. For example, if an entity has both the relation Org-Aff and Phys, then it cannot have the relation Art. But the GNN structure does not necessarily capture such complicated multiRoR that involve nested conditions. 

Therefore, we need another module to directly model the relation matrix $\bm{R}\in \mathbb{R}^{M\times M}$, which takes into consideration the dynamics among all relations as a whole, in order to capture multiRoR.

To this end, we propose a simple but effective module, a relation matrix transformer. As each relation $r_{ij}$ in the relation matrix $\bm{R}$ needs to attend to all other relations, we build our relation matrix transformer by customizing the Transformer encoder architecture, which allows extensive mutual attention among all elements \cite{DBLP:conf/nips/VaswaniSPUJGKP17}. Specifically, we customize the position encoding in the Transformer into two parts: row encoding and column encoding, each of which is a learnable mapping from the position index to a $d$-dimensional vector space. 

Our relation matrix transformer adds the position encoding, namely the sum of the row and column embedding, to the initial representations of relations obtained by procedures in Section~\ref{sec:method_init}. The input is thus a tensor $\bm{T} \in \mathbb{R}^{M \times M \times d}$, where $d$ is the dimension of features in the embedding.
The matrix transformer then learns the dynamics among all relations as a whole, and outputs new features of all relations by a transformed matrix $\bm{T_{\mathrm{Transformed}}} \in \mathbb{R}^{M \times M \times d'}$ which captures the multiRoR. 

Finally, we add together the relation embeddings learned by the GNN and the matrix transformer, and feed them into the final classification layer to obtain the type of each relation.

\section{Experiments}
\subsection{Datasets}
We use two benchmark RE datasets to evaluate the performance of our model.\footnote{Since our focus is RE, we do not use a newly published RE dataset, DocRED \cite{yao2019docred}. It contains entity linking annotations which can be a multi-tasking objective, and turns the task into a different setting.}

\paragraph{ACE05} 
The ACE 2005 Multilingual Training Corpus (ACE05)~\cite{walker2006ace} is a standard RE dataset. It has six relation categories, detailed in Section~\ref{sec:ror_stats}. We process and split the dataset following the practice in \cite{DBLP:conf/emnlp/GormleyYD15,DBLP:conf/acl/PlankM13}.\footnote{\url{https://github.com/mgormley/ace-data-prep}} There are six subdomains in the dataset: Broadcast Conversations (BC), Broadcast News (BN), Conversational Telephone Speech (CTS), Newswire (NW), Usenet Newsgroups (UN), and Weblogs (WL). The statistics of the resulted dataset is in Table~\ref{tab:ace05_stats}.
The macro F1-score is the primary evaluation metric of the data, as used by most works. As ``no-relation'' is regarded as the negative label, all non-negative relations on wrong entities as treated as false positives. 

\begin{table}[!h]
  \centering
    \begin{tabular}{clc}
    \toprule
    & \textbf{Domain} & \textbf{\# Relations} \\
    \midrule
    \multirow{1}{*}{\textbf{Train}} &
    NW+BN &  34,669  \\
    \midrule
    \multirow{1}{*}{\textbf{Valid}} & BC\_dev & 5,717 \\
    \midrule
    \multirow{3}{*}{\textbf{Test}} & BC\_test & 6,692 \\
    & CTS & 11,683 \\
    & WL &  11,022 \\
    \bottomrule
    \end{tabular}%
  \caption{Statistics of ACE05 dataset. 
  }\label{tab:ace05_stats}
\end{table}
\paragraph{SemEval2018} 
We use SemEval 2018 Task 7.2 \cite{DBLP:conf/semeval/GaborBSQZC18} as the second RE dataset to evaluate our model.\footnote{We choose Task 7.2 instead of the Task 7.1 used by \citet{DBLP:conf/acl/WangTYCWXGP19} because Task 7.1 only tests the \textit{classification} of positive relations, but Task 7.2 is the standard relation \textit{extraction} which tests on both positive and negative relations.}
The corpus is collected from abstracts and introductions of
scientific papers, and there are six types of semantic relations in total.
Note that there are three subtasks of it: Subtask 1.1 and Subtask 1.2 are relation \emph{classification} on clean and noisy data, respectively; Subtask 2 is the standard relation \textit{extraction}, where the same training set as Subtask 1 is used, but the evaluation is to identify all relations including ``no-relation.''
Following the main systems~\cite{DBLP:conf/semeval/RotsztejnHZ18,DBLP:conf/semeval/NooralahzadehOL18}, we combine the training data of both Subtask 1.1 and 1.2 as the training data.
The dataset consists of 136,965 train, and 27,316 test examples. We count the relation of every entity pair as a sample, and if no specific relation is annotated for some entity pairs, we use ``no-relation'' as the label. The standard evaluation metric is also macro F1, which is used for the official ranking of submissions.

\subsection{Baselines for ACE05}
For ACE05, we compare our model with the following systems.

\paragraph{$\text{BERT}_\text{EntEmb}$}

We use the baseline from  \cite{DBLP:conf/acl/WangTYCWXGP19} (called ``$\text{BERT}_\text{sp}$ with entity-indicator on input-layer'' in the original paper), which is essentially a BERT with entity embedding on the input layer.

\paragraph{$\text{OnePass}_\text{MRE}$ and $\text{OnePass}_\text{SRE}$} 
We also compare with the state-of-the-art systems on ACE05 -- \text{OnePass} by \citet{DBLP:conf/acl/WangTYCWXGP19}. OnePass has two variations: single relation extraction ($\text{OnePass}_\text{SRE}$) that treats 
every single relation extraction as an independent classification task including separate encoding and classification, and multiple relation extraction ($\text{OnePass}_\text{MRE}$) that encodes all entities in one pass, and then do the pair-wises classification per relation. 

\paragraph{Other models} 
We also compare our model with other previous RE models such as \cite{DBLP:conf/emnlp/GormleyYD15,DBLP:conf/naacl/NguyenG15,DBLP:conf/ijcnlp/FuNMG17,DBLP:conf/emnlp/ShiFHZJLH18}.

\subsection{Baselines for SemEval 2018 Task 7.2}
For SemEval2018, we compare our results with the top 3 systems on the leaderboard. For a fair comparison, we compare our models against the non-ensemble and ensemble models separately. 

\paragraph{Ensemble models} The top 1 system \cite{DBLP:conf/semeval/RotsztejnHZ18} is an ensemble of CNNs and RNNs. Its training uses an ensemble size of 20, data augmentation, multi-task learning, and many other meticulous designs.
Notably, when implementing our models, we only use a small ensemble size of 5, with no further tricks.

\begin{table*}[!ht]
  \centering
    \begin{tabular}{l|c|cccc}
    \toprule
    \textbf{} & \multicolumn{1}{c|}{\textbf{Dev}} & \multicolumn{4}{c}{\textbf{Test}} \\
    \textbf{} & BC\_dev & BC\_test & CTS   & WL & \textbf{Overall} \\
    \hline
    \textbf{HybridFCM \cite{DBLP:conf/emnlp/GormleyYD15}} & -- & 63.48 & 56.12 & 55.17 & 58.26
    \\
    \textbf{DAN \cite{DBLP:conf/ijcnlp/FuNMG17}} & -- & 65.16 & 55.55 & 57.19 & 59.30
    \\
    \textbf{GSN \cite{DBLP:conf/emnlp/ShiFHZJLH18}} & -- & 66.38 & 57.92 & 56.84 & 60.38
    \\
    \textbf{$\text{BERT}_\text{EntEmb}$ \cite{DBLP:conf/acl/WangTYCWXGP19}} & 65.32 & 66.86 & 57.65 & 53.56 & 59.36 \\
    \textbf{$\text{OnePass}_\text{MRE}$ \cite{DBLP:conf/acl/WangTYCWXGP19}} & 67.46 & 69.25 & 61.7  & 58.48 & 63.14 \\
    \textbf{$\text{OnePass}_\text{SRE}$ \cite{DBLP:conf/acl/WangTYCWXGP19}} & 68.90 & \textbf{69.76} & 63.71  & 57.20 & 63.14 \\
    \hline
    \textbf{$\text{\modelname{}}_\text{base}$} & 69.95 & 66.51 & 61.84 & 57.67 &  62.01 \\
    \textbf{$\text{\modelname{}}_\text{bi-only}$} & 68.50 & \textbf{68.86} & 64.18 & 59.57 & 64.20 \\
    \textbf{$\text{\modelname{}}_\text{multi-only}$} & \textbf{71.64}    & 66.77    & 59.90    & 57.62 & 61.43 \\
    \textbf{$\text{\modelname{}}_\text{full}$}  & 70.59 & 68.63 & \textbf{64.49} & \textbf{59.67} & \textbf{64.26} \\
    \bottomrule
    \end{tabular}%
  \caption{Macro F1 on ACE05. Following \cite{DBLP:conf/acl/WangTYCWXGP19}, we reported the performances on the development set, and all subsets of the test set. ``Overall'' is the average of all three subsets in the test set. }\label{tab:ace05}
\end{table*}

\paragraph{Non-ensemble models} The 2nd model  \cite{DBLP:conf/semeval/LuanOH18} uses LSTMs to encode both word sequences and dependency
tree structures, and perform relation extraction between concepts on top of them. And the 3rd model \cite{DBLP:conf/semeval/NooralahzadehOL18} uses a CNN model over the shortest dependency path between two entities.

\subsection{Our Models}
We investigate the following four settings of our model.

\paragraph{$\text{\modelname{}}_\text{base}$} For the base model in our architecture, we use a customized version of the $\text{BERT}_\text{EntEmb}$ baseline. Instead of the extra embedding for entities in $\text{BERT}_\text{EntEmb}$, we use the sentence index to indicate entities, so that the pretrained properties of BERT can be maintained as much as possible. We use this model as the base model that we build our biRoR and multiRoR learning modules on.

\paragraph{$\modelname{}_\text{bi-only}$ and $\modelname{}_\text{multi-only}$}
Based on $\text{\modelname{}}_\text{base}$, we implement {$\modelname{}_\text{bi-only}$}, which only the biRoR module based on GNN, and {$\modelname{}_\text{multi-only}$}, which uses only the multiRoR module based on a relation matrix transformer. 

\paragraph{$\modelname{}_\text{full}$} Finally, we implement the full model $\modelname{}_\text{full}$ which adopts both the GNN and the relation matrix transformer to learn all types of RoR.


\subsection{Implementation details}
We follow \cite{DBLP:conf/acl/WangTYCWXGP19}'s practice to use BERT-base (uncased) as the pretrained model.  
The GNN and matrix transformer has 4 layers, 8-head attention, and the hidden size of attention layers is 512. The feedforward layers of GNN have 1024 hidden units, and that of matrix transformer has 4096 hidden units. The batch size is 8.  The learning rate is set to 5e-5 on ACE05 and 1e-4 on SemEval2018, and the warmup is 0.1. We use the Adam optimizer and the cosine scheduler of the python package Transformers.\footnote{\url{https://github.com/huggingface/transformers}} We train the model for 30 epochs on both datasets. 
We run all experiments with one Tesla V100 card on a Ubuntu system.


\section{Results and analysis}
\subsection{Main results}
\paragraph{ACE05}
We first analyze the experiments on the larger dataset, ACE05. From the main results in Table~\ref{tab:ace05}, we can see that in the last column (overall performance), our full model $\text{\modelname{}}_\text{full}$ achieves the strongest performance, outperforming all previous models. Specifically, our model improves over the strongest setting of OnePass by 1.12\%, which is a large margin on the ACE05 dataset. Our full model Moreover, on the subsets of the test set, such as CTS and WL, which is different from the BC domain seen in the development set, our model also demonstrate consistent improvement over the previous best model.
\begin{table}[ht]
  \centering
    \begin{tabular}{lc}
    \toprule
    \textbf{Model}    & \textbf{Macro} \\
    \midrule
    \multicolumn{2}{c}{\textbf{\emph{Non-ensemble models}}} \\
           \cite{DBLP:conf/semeval/LuanOH18} &  39.10  \\
           \cite{DBLP:conf/semeval/NooralahzadehOL18} &  33.60    \\
           \textbf{$\text{\modelname{}}_\text{base}$} & 38.83  \\
           \textbf{$\text{\modelname{}}_\text{bi-only}$}  & 41.99  \\
           \textbf{$\text{\modelname{}}_\text{multi-only}$} & 42.87  \\
           \textbf{$\text{\modelname{}}_\text{full}$} & \textbf{45.46}  \\
    \midrule
    \multicolumn{2}{c}{\textbf{\emph{Ensemble models}}} \\
           \cite{DBLP:conf/semeval/RotsztejnHZ18} & 49.30   \\
           \textbf{E-$\text{\modelname{}}_\text{base}$} & 46.47  \\
           \textbf{E-$\text{\modelname{}}_\text{bi-only}$}  & 51.50   \\
           \textbf{E-$\text{\modelname{}}_\text{multi-only}$} & \textbf{51.85}  \\
           \textbf{E-$\text{\modelname{}}_\text{full}$} & 51.56  \\
    \bottomrule
    \end{tabular}
    \caption{Macro F1 scores on SemEval 2018 Task 7.2.}

  \label{tab:res_semeval}%
\end{table}%

\paragraph{SemEval2018}

In Table~\ref{tab:res_semeval}, we compare our system with the top 3 systems \cite{DBLP:conf/semeval/RotsztejnHZ18,DBLP:conf/semeval/LuanOH18,DBLP:conf/semeval/NooralahzadehOL18} on the SemEval 2018 Task 7.2 Leaderboard. For the non-ensemble setting, our model is +6.26\% higher in performance than the most competitive models on the leaderboard. For the more difficult ensemble model leaderboard, we use a simpler setting (i.e., smaller ensemble size, and no data augmentation), but surpasses the top model by +2.55\%. The improvement shows that our model with no complicated designs can demonstrate strong performance. 

\paragraph{Summary of main experiments}
Experiments on both datasets validate the effectiveness of our new perspective into RE. Because we model all the relation extraction tasks of the same text together, our method absorbs more knowledge from the same data than other models. Therefore, without data augmentation for SemEval2018, and without domain adaptation on ACE05 \cite{DBLP:conf/ijcnlp/FuNMG17,DBLP:conf/emnlp/ShiFHZJLH18}, our model still surpasses all systems that utilizes extra data.

\subsection{Does the model learn multi-relation instances well?}

We want to further analyze our model's ability to learn complicated multiRoR relations, especially for some entities with a high number of non-empty relations. For the model outputs on the ACE05 test set, we look into the subset of relations centered on entities with $\geq2$ relations. On this subset, the macro F1 of the previous model OnePass is 61.19\%, which drops 2 percent under its overall reported performance. However, our method achieves 63.55\%, keeping almost the same performance as the reported 64.26\% on the whole test set. On the more challenging subset involving entities with $\geq3$ relations, our model performs 56.63\%, almost 4 percent over OnePass's F1 of 52.82\%. This shows that the advantage of our model is stronger than others as the relation prediction gets more challenging.

\subsection{Does the model learn the conditional probability of relations well?}

We also compare our model's ability to learn biRoR versus OnePass. Remember the ground-truth conditional probability of every two relations mentioned before in Figure~\ref{fig:corr2} of Section~\ref{sec:corr1c}. For each row of the heatmap, it is normalized to probability 1, so each value in the row forms the probability distribution of any other relation conditioned on the observation of an existing relation. On the overall test set of ACE05, we analyze the output of our model and OnePass to obtain the corresponding heatmap, with each row representing a probability distribution. We calculate the average Jenson-Shannon (JS) distance of all the distributions. We find that OnePass is 0.0805 away from the gold distribution, and we are 0.0326 away from the gold distribution, $60\%$ closer than OnePass, which is the strongest previous model.

\section{Related Work}
Relation Extraction is one of the most important tasks in NLP. Conventional classification
approaches have made use of contextual, lexical
and syntactic features combined with richer linguistic and background knowledge such as WordNet and FrameNet \cite{DBLP:conf/semeval/HendrickxKKNSPP10,DBLP:conf/semeval/RinkH10}, as well as kernel-based methods \cite{DBLP:journals/jmlr/ZelenkoAR03,DBLP:conf/naacl/BunescuM05,DBLP:conf/acl/ZhouSZZ05}.

The recent advancement of deep neural networks result in a revolution in the methodology of RE. Many CNN-based \cite{DBLP:conf/coling/ZengLLZZ14,DBLP:conf/acl/SantosXZ15,DBLP:conf/naacl/NguyenG15}, and RNN-based \cite{DBLP:conf/emnlp/SocherHMN12,DBLP:journals/corr/ZhangW15a,DBLP:conf/acl/MiwaB16,DBLP:conf/acl/ZhouSTQLHX16} models achieve high performance in many datasets. The popularity of the field also gives birth to many shared tasks \cite{DBLP:conf/semeval/HendrickxKKNSPP10} which turned into the cradle of many competitive, well-designed systems \cite{DBLP:conf/semeval/LuanOH18,DBLP:conf/semeval/RotsztejnHZ18,DBLP:conf/semeval/JinDSMC18}.
Recently, as the model innovation in single relation extraction gradually slow down, most work shifted to the direction of distant supervision \cite{DBLP:conf/acl/MintzBSJ09,DBLP:conf/emnlp/ZengLC015,DBLP:conf/acl/LinSLLS16}.

These data-augmentation methods through distant supervision are orthogonal to the innovation in supervised models, and the focus of this paper is to innovate the supervised models to learn the same data with more thorough exploitation. To the best of our knowledge, we are the first paper in RE aiming to learn the correlation of relations in the given data.

In terms of our matrix formulation of the RoR, the most similar work is the table-filling approach of joint entity and relation extraction \cite{miwa2014modeling,gupta2016table}. However, our basic unit is an entity as opposed to every word in the text. Moreover, our proposed GNN and Matrix Transformer are different from \citeauthor{miwa2014modeling}'s \citeyearpar{miwa2014modeling} history-based structured learning with complex features and heuristics, as well as \citeauthor{gupta2016table}'s \citeyearpar{gupta2016table} RNN sequence decoding model.


\section{Ethical Considerations}
As with any RE algorithms, there is the danger of using the model to analyze the massive amount of text online, and retrieve the relationship among users and mine their critical information. We are aware of this danger, and sought to minimize the risk. For this reason, we only work on anonymized data, and the set of relations only involves common knowledge graph relations such as part-whole. The algorithm provided in this paper should \textit{not} be used to analyze any user-sensitive data, but as a tool to facilitate public knowledge graph construction.

\section{Conclusion}
In this paper, we proposed a new paradigm of RE, which is capable of modeling the interdependency among multiple relations in the same text. Our model uses a GNN and a matrix transformer to capture the RoR in data. Experiments validated that our model has a substantial improvement on the two benchmark datasets where most models compete for improvement on the decimal point. We also conducted analyses to reflect on our model's performance on learning multi-relation data and the proximity of distributions of correlation of the gold and predicted relations.


\section*{Acknowledgments}
We appreciate Prof Rada Mihalcea and Di Jin for their constructive suggestions on the storyline of this paper. We also thank Qipeng Guo for advice on the design of GNNs. Zhijing Jin appreciates Yixuan Zhang and Zhutian Yang for their support.

\bibliography{anthology,emnlp-ijcnlp-2019.bib}
\bibliographystyle{acl_natbib}

\appendix
\section{Appendices}
\subsection{Full List of Incompatibility Rules}\label{appd:corr1a}
The 12 incompatibility rules of entity type-constrained biRoR introduced in Section~\ref{sec:corr1a} are
\begin{enumerate}
\item Part-Whole (arg0) and Per-Soc (arg0)
\item Part-Whole (arg0) and Gen-Aff (arg0)
\item Part-Whole (arg1) and Per-Soc (arg0)
\item Part-Whole (arg1) and Gen-Aff (arg0)
\item Per-Soc (arg0) and Org-Aff (arg1)
\item Per-Soc (arg0) and Art (arg1)
\item Org-Aff (arg0) and Art (arg1)
\item Org-Aff (arg1) and Art (arg1)
\item Org-Aff (arg1) and Gen-Aff (arg0)
\item Art (arg0) and Art (arg1)
\item Art (arg1) and Gen-Aff (arg0)
\item Art (arg1) and Gen-Aff (arg1)
\end{enumerate}

\end{document}